\documentclass{article}
\usepackage{spconf,amsmath,amssymb,graphicx}
\usepackage{booktabs}
\usepackage{url}
\usepackage{balance}
\usepackage{hyperref}
\usepackage{xcolor}

\usepackage{orcidlink}

\makeatletter
\def\@ssect#1#2#3#4#5{%
  \begingroup\bf\centering
    {\interlinepenalty\@M \uppercase{#5}\par}%
  \endgroup
  \@tempskipa #3\relax
  \@xsect{\@tempskipa}}
\makeatother

\let\oldthebibliography\thebibliography
\renewcommand{\thebibliography}[1]{%
  \oldthebibliography{#1}%
  \setlength{\itemsep}{0.4pt}\setlength{\parskip}{0pt}\setlength{\parsep}{0pt}}

\title{GA-EIRFS: A Geometry-Augmented Repeat-Factor Sampling Method \\ for
Long-Tailed LiDAR 3D Object Detection}

\name{\begin{tabular}{c}
Taufiq Ahmed$^{\dagger}$$^{\orcidlink{0009-0002-0193-8781}}$, Constantino {\'A}lvarez Casado$^{\dagger}$$^{\orcidlink{0000-0002-3052-4759}}$, Daniel Herrera Castro$^{\dagger}$$^{\orcidlink{0009-0009-8320-5530}}$,\\
Sasan Sharifipour$^{\dagger}$$^{\orcidlink{0000-0001-6670-4252}}$, Abhishek Kumar$^{\star}$$^{\orcidlink{0000-0003-4383-7225}}$, Miguel Bordallo L{\'o}pez$^{\dagger}$$^{\orcidlink{0000-0002-5707-9085}}$
\end{tabular}}

\address{
$^{\dagger}$Center for Machine Vision and Signal Analysis (CMVS),
University of Oulu, Finland \\
$^{\star}$University of Jyv\"{a}skyl\"{a}, Jyv\"{a}skyl\"{a}, Finland \\
}

\begin{document}
\ninept
\maketitle

\begin{abstract}
Long-tailed 3D object detection is treated as a class-frequency problem, but LiDAR supervision quality depends on object observability: similar frequencies can hide different geometric evidence. We introduce Geometry-Augmented Exponentially Weighted Instance-Aware Repeat Factor Sampling (GA-EIRFS), a detector-agnostic method that modulates a frequency-based repeat factor with a fixed geometry score combining point count, surface-normal entropy, and surface coverage. GA-EIRFS changes only frame-sampling probabilities, leaving the detector and inference unchanged. On nuScenes it improves mean average precision (mAP) and the nuScenes detection score (NDS) in four converged experiments with CenterPoint and PointPillars over two seeds; for CenterPoint at seed 666, mAP rises from 0.552 to 0.563 and bicycle AP from 0.306 to 0.359. Per-class gains correlate with the class sampling-weight increase (Spearman $\rho=0.70$, $p=0.025$) but not with geometry score alone ($\rho=0.32$, $p=0.37$), so geometry amplifies frequency-driven need. KITTI results vary across seeds, most for the rarest class. Code: \href{https://github.com/Multimodal-Sensing-Lab/GA-EIRFS}{https://github.com/Multimodal-Sensing-Lab/GA-EIRFS}.

\end{abstract}


%
%
\section{Introduction}
\label{sec:intro}

LiDAR-based 3D object detection supports \textcolor{black}{autonomous vehicles, mobile robots, inspection drones and roadside sensing, where missing an uncommon but safety-relevant object can have serious consequences}. Annotated datasets often contain far fewer examples of such objects than of common categories. For example, in the nuScenes database training split~\cite{Caesar2020nuScenes}, annotated cars outnumber bicycles by 41.5:1, which presents a challenge in learning to detect underrepresented classes~\cite{Peri2022Towards,Lee2025NBA3D}. \textcolor{black}{We evaluate on driving benchmarks, but the underlying problem of imbalanced training data extends to other LiDAR applications}.



Sampling-based rebalancing addresses the imbalance without modifying the detector. Repeat Factor Sampling (RFS)~\cite{gupta2019lvis} repeats an image according to the frequency of the rarest category it contains. Instance-Aware RFS (IRFS)~\cite{yaman2023instance} combines image and instance frequency through a geometric mean, and E-IRFS~\cite{Ahmed2025EIRFS} applies an exponential function to that mean so that extremely rare categories receive a stronger adjustment. In 3D detection, Class-Balanced Grouping and Sampling (CBGS) duplicates frames by category occurrence~\cite{Zhu2019CBGS}. Every method in this family reads annotation counts only.

\begin{figure}[t]
    \centering
    \includegraphics[width=0.94\linewidth]{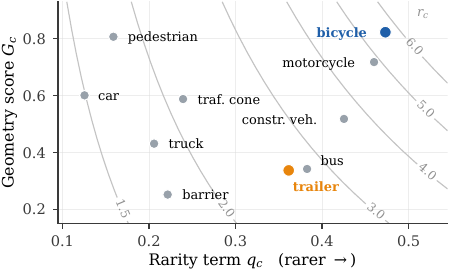}
    \vspace{-3mm}
    \caption{Rarity and geometric difficulty are close to independent across the ten nuScenes classes. Markers are classes, placed by the E-IRFS frequency term $q_c$ and the geometry score $G_c$ of Eq.~\eqref{eq:geometry}. Grey curves are contours of $r_c$ ($\alpha=2.0$, $\beta=1.0$): vertical distance at fixed rarity is the \textcolor{black}{sampling weight} that geometry adds.}
    \label{fig:motivation}
\end{figure}

Frequency is not the same as observability. LiDAR returns depend on object size, range, orientation, occlusion and surface structure. In the nuScenes training split a bicycle box contains 16.9 points on average, against 98.1 for a car and 240.3 for a trailer. Bicycle and trailer are both minority classes and a frequency-based rule gives them similar \textcolor{black}{sampling weight}, yet the evidence available per instance differs by more than an order of magnitude. Across the ten classes the frequency term and the geometry score are close to uncorrelated (Spearman $\rho=0.16$, $p=0.65$), as Figure~\ref{fig:motivation} shows, so a rule defined on counts alone cannot separate a rare class that is easy to observe from one that is not.


We propose Geometry-Augmented E-IRFS (GA-EIRFS), which estimates a fixed class-level geometry score from the annotated LiDAR points before training and uses it to modulate the exponent of the E-IRFS repeat factor. Geometry amplifies an existing frequency-driven adjustment rather than acting on its own, and one parameter controls how much it contributes. The contributions are as follows: 

\begin{itemize}
\setlength\itemsep{0.5pt}
\item A geometry-augmented repeat factor that combines class frequency with class-level LiDAR observability and reduces exactly to E-IRFS when the geometry term is disabled.
\item A geometry score built from three measurable properties of the annotated point sets, computed once before training, with no change to the detector, the loss or the inference path.
\item An evaluation on nuScenes with two detectors and two seeds, a sweep of the geometry strength, an analysis of which classes benefit and why, and a transfer experiment on KITTI in which the gains are not consistent.
\end{itemize}

%
%
\section{Related Work}
\label{sec:related}

\noindent\textbf{Rebalancing for long-tailed 3D detection.}
Long-tail mitigation for point clouds acts on the data distribution, on the training objective, or on the detector. CBGS builds a more balanced training distribution and groups related categories~\cite{Zhu2019CBGS}, Lee and Kim combine category-specific heads, dynamic loss averaging and contextual ground-truth sampling~\cite{Lee2023Resolving}, and Peri \emph{et al.} formalize long-tailed 3D detection with hierarchical supervision and evaluation~\cite{Peri2022Towards}. Later work adds multimodal fusion~\cite{ma2023long}, neighbour-based confidence adjustment~\cite{Lee2025NBA3D}, loss reweighting~\cite{Chang2024Optimizing} or few-shot adaptation~\cite{Liu2023Generalized}. These methods change the objective, the architecture or the modalities\textcolor{black}{, and none of them changes which frames the sampler draws}. Frequency and difficulty have been separated before, but outside 3D geometry: CDB-S resamples classes by measured validation difficulty~\cite{Sinha2022Class} and geometric priors have shaped the feature space in long-tailed classification~\cite{Ma2024Geometric}. Both derive difficulty from model behaviour rather than from the sensor evidence.

\noindent\textbf{Geometry and difficulty-aware training.}
Point density has been used to guide internal point selection and attention in DA-3DSSD~\cite{Ning2021Density} and in density-aware set abstraction~\cite{Zhang2023Boosting}. PointDrop generates sparse adversarial examples~\cite{Ma2021PointDrop}, and density-adaptive augmentation modifies point-cloud content~\cite{Lee2023Dual}. All of these act inside the detector or inside a scene. The closest operational method is Curricular Object Manipulation (COM)~\cite{Zhu2023Curricular}, which groups database objects by distance, dimensions, orientation and occupancy, then progressively selects harder objects for ground-truth copy-paste augmentation. Its sampling unit is an object pasted into a scene and its difficulty is updated from model scores during training. Rare Example Mining~\cite{Jiang2022Improving} also separates rareness from difficulty, but it selects unlabelled tracks for annotation rather than rebalancing a labelled set, as do active-annotation methods~\cite{Feng2019Deep, 10158738}.

\noindent\textcolor{black}{\textbf{Gap and positioning.} Two gaps follow. The sampling methods decide how often a frame is drawn from annotation counts alone, so they cannot separate a rare class the sensor observes well from one it observes poorly. The geometry-aware methods do read sensor structure, but they act inside the detector or inside a synthesised scene, and those that separate rarity from difficulty estimate difficulty from model behaviour or select data for annotation. We found no method that inserts a fixed, sensor-derived geometry score directly into a frame repeat factor. GA-EIRFS targets that gap: it scores class-level observability once from the annotated points and multiplies it into the exponent of an existing frequency-based repeat factor, leaving the detector, the loss and inference untouched. Section~\ref{sec:method} defines the two priors and how they combine.}

%
%
\section{Geometry-Augmented Repeat Factor Sampling}
\label{sec:method}

\begin{figure}[ht!]
    \centering
    \includegraphics[width=\columnwidth]{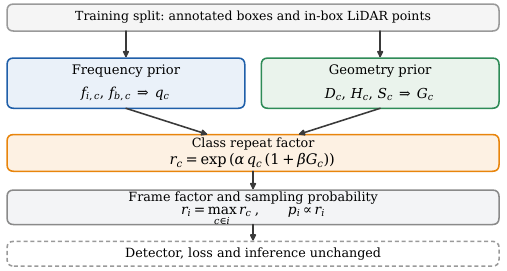}
    \vspace{-5mm}
    \caption{GA-EIRFS pipeline. The priors $q_c$ and $G_c$ of Eqs.~\eqref{eq:frequency} and~\eqref{eq:geometry} are estimated once from the training split and enter the class repeat factor of Eq.~\eqref{eq:gaeirfs}, which becomes a frame sampling probability.}
    \label{fig:pipeline}
\end{figure}

GA-EIRFS takes an annotated training split as input and returns one sampling probability per frame. Figure~\ref{fig:pipeline} shows the two priors, their combination, and the single point at which the training loop is affected.

\subsection{Frequency prior}

Let $f_{i,c}$ be the fraction of training frames containing class $c$ and $f_{b,c}$ be the fraction of 3D annotated boxes assigned to $c$. Following E-IRFS~\cite{Ahmed2025EIRFS}, the frequency term is given by:
\begin{equation}
q_c=\sqrt{\frac{t}{\sqrt{f_{i,c}\,f_{b,c}}}},
\label{eq:frequency}
\end{equation}
where $t$ sets the threshold at which oversampling becomes active. The term grows as either frequency falls and treats classes with comparable occurrence statistics identically.

\subsection{Class-level geometry prior}

For every annotated instance, we collect the LiDAR points inside its ground-truth 3D box. Let $\mathcal{N}(\cdot)$ denote min-max normalization across the classes of the dataset. Each component is constructed to lie in $[0,1]$, with higher values indicating a greater estimated geometric difficulty. The first component penalizes low point support and is defined as:
\begin{equation}
D_c=1-\mathcal{N}\!\left(\log(1+\bar{n}_c)\right),
\label{eq:density}
\end{equation}
where $\bar{n}_c$ is the mean number of in-box points per instance of class $c$. The logarithm reduces the influence of classes with very high point counts on the normalization. In nuScenes, the traffic cone class has $D_c=1.00$, with 9.2 points per instance on average, while the bus class has $D_c=0.00$, with 312.4 points per instance.

The second component measures surface diversity through the distribution of local surface-normal orientations. Normals are estimated for each instance and assigned to 256 orientation bins. The normalized class-level entropy is defined as:
\begin{equation}
H_c=\mathcal{N}\!\left(-\sum_{k=1}^{256}p_{c,k}\log p_{c,k}\right),
\label{eq:entropy}
\end{equation}
where $p_{c,k}$ is the probability assigned to orientation bin $k$ for class $c$. A flat panel exhibits similar normal orientations and low entropy, whereas an irregular structure such as a bicycle frame can exhibit more diverse orientations and higher entropy. Instances with fewer than ten points are excluded from the entropy calculation because their normal estimates are unstable. Assigning zero entropy to these instances would artificially lower the estimated surface diversity of sparsely sampled classes.

The third component measures how sparsely the observed surface is covered,
\begin{equation}
S_c=1-\mathcal{N}\!\left(\overline{\left(n/A\right)}_c\right),
\qquad A=2(lw+lh+wh),
\label{eq:coverage}
\end{equation}
where $n$ is the in-box point count and $A$ the surface area of a box of dimensions $l$, $w$ and $h$. The surface area replaces volume because a LiDAR sensor observes the outer surfaces, and normalizing by volume makes large hollow boxes such as the trailer and bus appear artificially empty. The geometry score is the convex combination
\begin{equation}
G_c=\lambda_D D_c+\lambda_H H_c+\lambda_S S_c,
\qquad \lambda_D+\lambda_H+\lambda_S=1 .
\label{eq:geometry}
\end{equation}

\subsection{Geometry-augmented repeat factor}

GA-EIRFS modulates the E-IRFS exponent with the geometry score computed as:
\begin{equation}
r_c=\exp\!\left(\alpha\, q_c\,(1+\beta G_c)\right),
\label{eq:gaeirfs}
\end{equation}
where $\alpha$ controls the exponential scaling and $\beta$ the geometry contribution. Setting $\beta=0$ recovers E-IRFS exactly, allowing direct ablation of the geometry term. For a frame $i$, the sampling weight is $r_i=\max_{c\in i} r_c$, and its probability of being selected for training is $p_i=r_i/\sum_j r_j$, where the sum runs over all training frames.


The product form in Eq.~\eqref{eq:gaeirfs} is deliberate. Geometry scales a \textcolor{black}{weight} that rarity has already requested, so a common class that happens to be geometrically difficult is not promoted on its own. Table~\ref{tab:data} reports the effect. Bicycle and trailer receive similar \textcolor{black}{sampling weight} under E-IRFS, at 2.58 and 2.06, a ratio of 1.25. Under GA-EIRFS the factors become 5.61 and 2.63 and the ratio widens to 2.14, while car moves only from 1.29 to 1.49. Taking the maximum over the classes in a frame, rather than a product, avoids compounding when several minority classes co-occur and preserves the frame-level behaviour of E-IRFS, so that $\beta$ isolates geometry. The cost is one offline pass over the annotated in-box points.


%
%
\vspace{-1mm}
\section{Experimental Setup}
\label{sec:setup}
\vspace{-1mm}
\noindent\textbf{Datasets.} The main evaluation uses nuScenes~\cite{Caesar2020nuScenes}, comprised of 1000 driving scenes recorded with a 32-beam LiDAR at 20\,Hz. We use the standard split of 700 training and 150 validation scenes, that is 28,130 and 6,019 annotated keyframes, and the ten official detection classes. Transfer is assessed on KITTI~\cite{Geiger2013KITTI}, which uses a 64-beam sensor, single sweeps and three classes, and whose Car to Cyclist ratio of about 19.6 to 1 is milder than the 41.5 to 1 ratio on nuScenes.


\noindent\textbf{Models, protocol and metrics.} We evaluate CenterPoint~\cite{Yin2021CenterPoint} and PointPillars~\cite{8954311} in OpenPCDet~\cite{openpcdet2020}. The principal nuScenes runs use 20 epochs and the seeds 666 and 1337, \textcolor{black}{that is two repetitions per detector and four paired baseline-to-GA-EIRFS comparisons in total}. We set $\alpha=2.0$, $t=0.01$, $\beta=1.0$ and $(\lambda_D,\lambda_H,\lambda_S)=(0.5,0.3,0.2)$, weighting density most because it is measured directly rather than estimated. The geometry sweep uses 12 epochs and $\beta\in\{0,0.5,1.0,2.0\}$. For KITTI we keep $\alpha$, $t$ and $\beta$ unchanged, recompute $G_c$ from the KITTI point clouds, use PointPillars with the reference OpenPCDet configuration and no road-plane augmentation, and evaluate the seeds 666, 1337 and 42 (\textcolor{black}{3 repetitions}). We report nuScenes \textcolor{black}{mean average precision (mAP) and the nuScenes detection score (NDS)}: mAP matches \textcolor{black}{predictions to ground truth} by center distance, \textcolor{black}{while NDS combines mAP with the true-positive errors in} translation, scale, orientation, velocity and attribute~\cite{Caesar2020nuScenes}; per-class AP; and KITTI moderate-difficulty 3D AP via box IoU~\cite{Geiger2013KITTI}. Every comparison is paired: baseline and GA-EIRFS share detector, schedule, augmentation and seed, and differ only in the sampler.


\begin{table}[ht!]

\caption{nuScenes class statistics and repeat factors, ordered by instance count. Eq.~\eqref{eq:gaeirfs} uses $\alpha=2$, $t=0.01$: $\beta=0$ gives E-IRFS and $\beta=1$ our default.}
\label{tab:data}
\centering
\setlength{\tabcolsep}{3.1pt}
\def\arraystretch{0.92}
\begin{tabular}{lrrrrr}
\toprule
 & & Points/ & & \multicolumn{2}{c}{Repeat factor $r_c$}\\
\cmidrule(lr){5-6}
Class & Instances & obj. & $G_c$ & $\beta{=}0$ & $\beta{=}1$\\
\midrule
Car                   & 339,949 &  98.1 & 0.601 & 1.29 & 1.49\\
Pedestrian            & 161,928 &  11.7 & 0.807 & 1.37 & 1.78\\
Barrier               & 107,507 &  62.6 & 0.252 & 1.56 & 1.74\\
Truck                 &  65,262 & 209.8 & 0.431 & 1.51 & 1.80\\
Traffic cone          &  62,964 &   9.2 & 0.588 & 1.61 & 2.14\\
\addlinespace[1.5pt]
Trailer               &  19,202 & 240.3 & 0.337 & 2.06 & 2.63\\
Bus                   &  12,286 & 312.4 & 0.342 & 2.15 & 2.79\\
Construction vehicle  &  11,050 & 103.9 & 0.518 & 2.34 & 3.64\\
Motorcycle            &   8,846 &  41.4 & 0.718 & 2.51 & 4.86\\
Bicycle               &   8,185 &  16.9 & 0.823 & 2.58 & 5.61\\
\bottomrule
\end{tabular}
\end{table}

\begin{figure*}[ht!]
    \centering
    \includegraphics[width=0.97\textwidth]{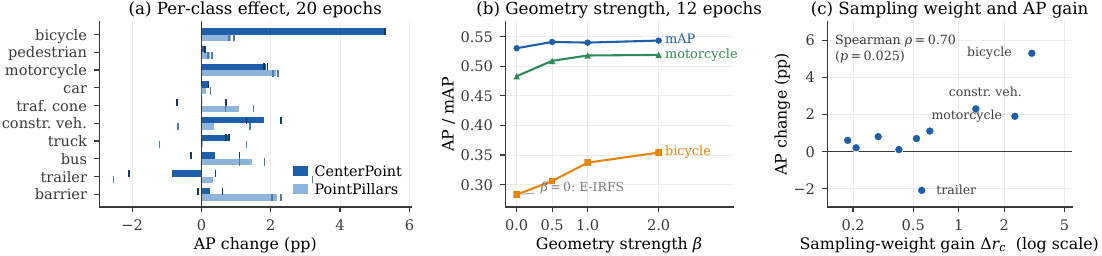}
    \vspace{-4mm}
    \caption{Quantitative evidence on nuScenes. (a) Per-class AP change at 20 epochs, in percentage points (pp). Bars are the mean of the two seeds, vertical markers are the individual seeds, and classes are ordered by $G_c$, lowest at the bottom. (b) Geometry-strength sweep at 12 epochs, CenterPoint, seed 666. (c) AP change against the exposure gain $\Delta r_c=r_c(\beta{=}1)-r_c(\beta{=}0)$, with the Spearman rank correlation over the ten classes.}
    \label{fig:results}
\end{figure*}

\begin{figure*}[ht!]
    \centering
    \includegraphics[width=0.98\textwidth]{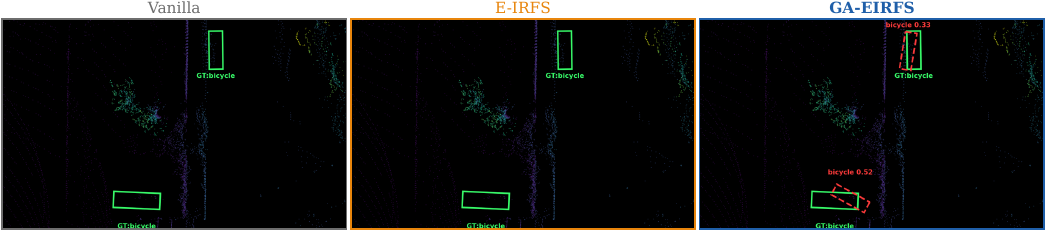}
    \vspace{-4mm}
    \caption{Top-down view of nuScenes validation sample 3896, in which the two annotated bicycles are far apart and sparsely sampled. Green boxes are ground truth, dashed red boxes are predictions scoring at least 0.3. Vanilla and E-IRFS return no bicycle above the threshold, whereas GA-EIRFS recovers both, at 0.33 and 0.52. The panels differ only in the sampler.}
    \label{fig:qual}
\end{figure*}

%
%
\vspace{-1mm}
\section{Results and Discussion}
\label{sec:results}
\vspace{-1mm}

\noindent\textbf{Overall performance.}
Table~\ref{tab:main} reports the four fully converged nuScenes comparisons. GA-EIRFS increases both mAP and NDS in every run, \textcolor{black}{by 0.4 to 1.3\,pp mAP and 0.3 to 0.7\,pp NDS. Over the four paired runs the mean gain is $+0.91$\,pp mAP (95\% confidence interval $[0.27,1.55]$, paired $t$ test $p=0.020$) and $+0.49$\,pp NDS ($[0.12,0.86]$, $p=0.025$). Both intervals exclude zero, although four observations pooled over two architectures bound the effect rather than fix it.} The two detectors differ in backbone, feature representation and detection head, so the agreement indicates that the effect is not tied to one architecture. The 12-epoch runs isolate the geometry term from the exponential frequency weighting. E-IRFS reproduces the vanilla mAP exactly under both seeds, 0.5303 and 0.5361, while adding geometry raises it to 0.5398 and 0.5428, so the improvement is attributable to $G_c$ and not to the reweighting it modulates.

\vspace{-3pt}

\begin{table}[ht!]
\def\arraystretch{1.0}
\setlength{\tabcolsep}{0.45em}
\caption{nuScenes validation results after 20 epochs. V is training without rebalancing and GA is GA-EIRFS. Higher is better, $\Delta$ is in percentage points computed before rounding, and the better value of each metric within a run is in bold. \textcolor{black}{The last row is the mean of the four paired differences, with 95\% confidence intervals $[+0.27,+1.55]$ for mAP and $[+0.12,+0.86]$ for NDS.}}
\label{tab:main}
\centering
\begin{tabular}{llrrrrrr}
\toprule
& & \multicolumn{3}{c}{mAP $\uparrow$} & \multicolumn{3}{c}{NDS $\uparrow$}\\
\cmidrule(lr){3-5}\cmidrule(lr){6-8}
Detector & Seed & V & GA & $\Delta$ & V & GA & $\Delta$\\
\midrule
CenterPoint  & 666  & 0.552 & \textbf{0.563} & $+1.1$ & 0.635 & \textbf{0.642} & $+0.7$\\
CenterPoint  & 1337 & 0.554 & \textbf{0.563} & $+0.9$ & 0.638 & \textbf{0.641} & $+0.3$\\
PointPillars & 666  & 0.385 & \textbf{0.388} & $+0.4$ & 0.536 & \textbf{0.538} & $+0.3$\\
PointPillars & 1337 & 0.382 & \textbf{0.395} & $+1.3$ & 0.534 & \textbf{0.541} & $+0.7$\\
\midrule
\multicolumn{2}{l}{\textcolor{black}{Mean, 4 runs}} & & & \textcolor{black}{$+0.9$} & & & \textcolor{black}{$+0.5$}\\
\bottomrule
\end{tabular}
\end{table}

\noindent\textbf{Which classes benefit.}
We divide the ten classes of Table~\ref{tab:data} into the 5 rarest and the 5 most frequent. For CenterPoint at seed 666 the mean AP of the rare group rises from 0.393 to 0.410, a gain of 1.7\,pp, whereas the frequent group moves from 0.712 to 0.716. Bicycle improves from 0.306 to 0.359, a gain of 5.3\,pp or 17\%. Figure~\ref{fig:results}(a) shows how consistently these effects repeat. Across the four runs, overall mAP and NDS improve in four of four, as do bicycle and motorcycle, while barrier, traffic cone, bus, truck and construction vehicle improve in three of four. Trailer splits two against two, so its behaviour is better described as run-to-run variation than as a systematic loss. Car and pedestrian, which already saturate, never degrade.


\noindent\textbf{Why they benefit.}
The geometry score alone does not predict the per-class outcome. Over the ten classes the rank correlation between $G_c$ and AP change is weak ($\rho=0.32$, $p=0.37$), whereas the \textcolor{black}{gain in the class sampling weight} $\Delta r_c=r_c(\beta{=}1)-r_c(\beta{=}0)$ correlates considerably better ($\rho=0.70$, $p=0.025$), as plotted in Figure~\ref{fig:results}(c). \textcolor{black}{This quantity measures a change in class sampling weight, not observed training exposure: frame weights are determined by the maximum class weight and normalized across all training frames, so actual class exposure also depends on co-occurrence.} This supports the mechanism the product form was designed for: geometry is useful when it amplifies a frequency-driven need for exposure and has little effect when the frequency term is already small. The sweep in Figure~\ref{fig:results}(b) gives the most direct evidence that $G_c$ drives the effect: bicycle AP rises monotonically from 0.283 at $\beta=0$ to 0.354 at $\beta=2$, motorcycle AP follows the same ordering, and overall mAP stays within 0.34\,pp for every nonzero $\beta$. A range-corrected $G_c$ preserves the class ranking ($\rho=0.939$), so the score is not merely a proxy for the distance distribution of a class. With ten class observations, these $p$-values are diagnostic rather than confirmatory.


\noindent\textbf{Qualitative behaviour.}
Figure~\ref{fig:qual} shows a validation frame with two bicycles at different ranges. Neither vanilla training nor E-IRFS produces a bicycle prediction above the operating threshold, while GA-EIRFS recovers both. The frame was selected as a case in which the frequency-only sampler and the baseline behave identically, the condition the geometry term is meant to change. One frame is not evidence of the size of the effect.

\noindent\textbf{Transfer to KITTI.}
On KITTI, Cyclist is the rarest class and, independently, has the highest geometry score at $G_c=0.935$, giving it the largest repeat factor at 4.05 against 1.92 for Pedestrian and 1.38 for Car, so the pattern that motivates the method on nuScenes reappears under a different sensor. The detection results do not follow as cleanly. Table~\ref{tab:kitti} shows that both minority classes lean positive, but neither improvement holds across the three seeds. Pedestrian gains 0.69\,pp on average, while Cyclist loses 0.16\,pp because the seed 1337 drop outweighs its two gains, and Car is unchanged. We treat KITTI as a boundary condition rather than a generalisation result. A likely contributor is pool size: Cyclist has 734 instances in 514 frames, the smallest pool in either dataset, so its large repeat factor concentrates exposure on few frames, the same mechanism behind trailer's run-to-run variation on nuScenes.

\begin{table}[ht!]
\def\arraystretch{1.0}
\setlength{\tabcolsep}{1.05em}
\caption{KITTI 3D AP (\%) at moderate difficulty, PointPillars, three seeds. Better value of each pair in bold.}
\label{tab:kitti}
\centering
\scriptsize
\begin{tabular}{lrrrrrr}
\toprule
& \multicolumn{2}{c}{Car} & \multicolumn{2}{c}{Pedestrian} & \multicolumn{2}{c}{Cyclist}\\
\cmidrule(lr){2-3}\cmidrule(lr){4-5}\cmidrule(lr){6-7}
Seed & V & GA & V & GA & V & GA\\
\midrule
666   & \textbf{75.88} & 75.87 & 44.19 & \textbf{45.34} & 61.90 & \textbf{62.29}\\
1337  & \textbf{76.28} & 75.81 & 41.93 & \textbf{43.74} & \textbf{62.69} & 59.99\\
42    & 75.34 & \textbf{75.61} & \textbf{44.89} & 44.00 & 61.05 & \textbf{62.88}\\
\midrule
Mean  & \textbf{75.83} & 75.76 & 43.67 & \textbf{44.36} & \textbf{61.88} & 61.72\\
\textcolor{black}{SD}   & \textcolor{black}{0.47} & \textcolor{black}{0.14} & \textcolor{black}{1.55} & \textcolor{black}{0.86} & \textcolor{black}{0.82} & \textcolor{black}{1.53}\\
\bottomrule
\end{tabular}
\end{table}

\noindent\textbf{Limitations.}
GA-EIRFS uses hand-set mixture weights and one score per class. It cannot represent within-class variation caused by range, weather or occlusion, and repeating a frame also repeats the frequent objects and background it contains, which is the most likely reason why classes such as trailer respond inconsistently. The nuScenes evidence covers two detectors and two seeds, so the associations constrain the mechanism but do not establish it, and the method was not compared directly against curriculum-based augmentation under a shared protocol.

%
%
\section{Conclusion}
\label{sec:conclusion}

We presented GA-EIRFS, a rebalancing method that inserts a fixed class-level LiDAR geometry prior into a frequency-based frame repeat factor. On nuScenes, mAP and NDS increased in all four converged detector and seed combinations, bicycle AP increased by 5.3\,pp for CenterPoint at seed 666, and it grew monotonically with $\beta$. The gains track the realised sampling-weight change rather than the geometry score itself, which supports reading the method as an amplifier of frequency-driven sampling rather than as an independent difficulty sampler. The evidence is limited to two seeds per detector and to one dataset with a severe imbalance, since the KITTI transfer was not consistent across seeds. Learning the mixture weights and conditioning the score on range or on the instance are the next steps.

\section*{Acknowledgment}
The research was supported by the Business Finland WiSeCom project (Grant 3630/31/2024), the University of Oulu, the Research Council of Finland 6G Flagship Programme (Grant 346208), and the Business Finland 6GSoft project (Grant 8541/31/2022). Research supported by the NVIDIA Academic Grant Program using nVidia RTX6000 Blackwell GPUs.

\balance

\bibliographystyle{IEEEbib}
\bibliography{ga_eirfs_references}

\end{document}